\documentclass[
]{ceurart}

\usepackage{listings}
\usepackage{booktabs}
\usepackage{color, soul}
\usepackage{graphicx}
\usepackage{tikz}
\usepackage{pgfplots}
\usepackage{pgfplotstable}
\pgfplotsset{compat=newest}
\usepackage{tabu}
\usepackage{latexsym}
\usepackage[T1]{fontenc}
\usepackage[utf8]{inputenc}
\usepackage{microtype}
\usepackage{inconsolata}
\usepackage{float}
\usepackage{multirow}
\usepackage{multicol}
\usepackage{subfig}
\usepackage{hyperref}
\usepackage{subcaption}
\usepackage{wrapfig}
\usepackage{xurl}

\definecolor{darkblue}{rgb}{0, 0, 0.5}
\definecolor{darkorange}{rgb}{.9, .5, .1}

\begin{document}

\copyrightyear{2024}
\copyrightclause{Copyright for this paper by its authors.
  Use permitted under Creative Commons License Attribution 4.0
  International (CC BY 4.0).}

\conference{CAMLIS'24: Conference on Applied Machine Learning for Information Security,
  October 24--25, 2024, Arlington, VA}

\title{AdapterSwap: Continuous Training of LLMs with Data Removal and Access-Control Guarantees}

\author{William Fleshman}[%
email=will.fleshman@jhu.edu,
url=https://fleshman.dev,
]
\cormark[1]
\address{Johns Hopkins University,
  3101 Wyman Park Dr, Baltimore, MD 21218}

\author{Aleem Khan}
\author{Marc Marone}
\author{Benjamin {Van Durme}}

\cortext[1]{Corresponding author.}

\begin{abstract}
Large language models (LLMs) are increasingly capable of completing knowledge intensive tasks by recalling information from a static pretraining corpus. Here we are concerned with LLMs in the context of evolving data requirements.  For instance: batches of new data that are introduced periodically; subsets of data with user-based access controls; or requirements on dynamic removal of documents with guarantees that associated knowledge cannot be recalled. We wish to satisfy these requirements while at the same time ensuring a model does not forget old information when new data becomes available. To address these issues, we introduce AdapterSwap, a training and inference scheme that organizes knowledge from a data collection into a set of dynamically composed low-rank adapters. Our experiments demonstrate AdapterSwap's ability to support efficient continual learning, while also enabling organizations to have fine-grained control over data access and deletion.
\end{abstract}

\begin{keywords}
  LLM \sep
  adapter \sep
  access-control \sep
  data removal \sep
  forgetting
\end{keywords}

\maketitle

\section{Introduction}

Generative Large Language Models (LLMs) continue to improve in handling a broad range of Natural Language Understanding (NLU) and Natural Language Generation (NLG) tasks. To achieve these improvements, models have grown in size such that training or fine-tuning a full model for a custom task or data distribution is not possible on commodity hardware \footnote{For example, the Falcon-7B model used in our experiments was pretrained with 384 40GB A100 GPUs.}. Under such constraints, organizations may look to other solutions for leveraging LLMs with their own data. 

\emph{Parameter Efficient Fine-Tuning} (PEFT) is a collection of approaches for adapting a model to new tasks or data domains. One method of PEFT is to train a small number of new parameters (adapters) which enable the model to perform well in the current setting \cite{bapna-firat-2019-simple, houlsby2019parameterefficient, DBLP:journals/corr/abs-2106-09685}. These approaches allow organizations to update models with their data in a computationally efficient manner.

Beyond compute, organizations may face additional challenges such as incorporating knowledge from a continuous stream of data or ensuring a model adheres to user-based access-controls applied to the data. Additionally, data protection policies or legal outcomes could lead to organizations losing access to a subset of data, preventing the use of models which used that subset during training \cite{GDPR2016a, NYT2024}.

Accordingly, we introduce AdapterSwap, a parameter efficient approach for continual learning which addresses data access-control and removal while retaining the ability to acquire new knowledge. We achieve these results through a thoughtful consideration of Low-Rank Adaptation (LoRA) \cite{DBLP:journals/corr/abs-2106-09685}. First, data is segmented into appropriately sized groups based on access-control levels and available computing resources. A general purpose LM is used as a base model, with a LoRA adapter fine-tuned on each group. During inference, a retriever model is used to select \emph{adapters} relevant to the query, which are optionally filtered according to salient access-controls. A weighted combination of these adapters are then applied to the base model to produce an appropriate response. Crucially, if a document must later be removed from the corpus, only the impacted adapter will need to be retrained. From the user perspective, removing data is then a relatively low latency, computationally efficient process. 

\begin{figure}[ht]
    \centering
    \includegraphics[scale=0.58]{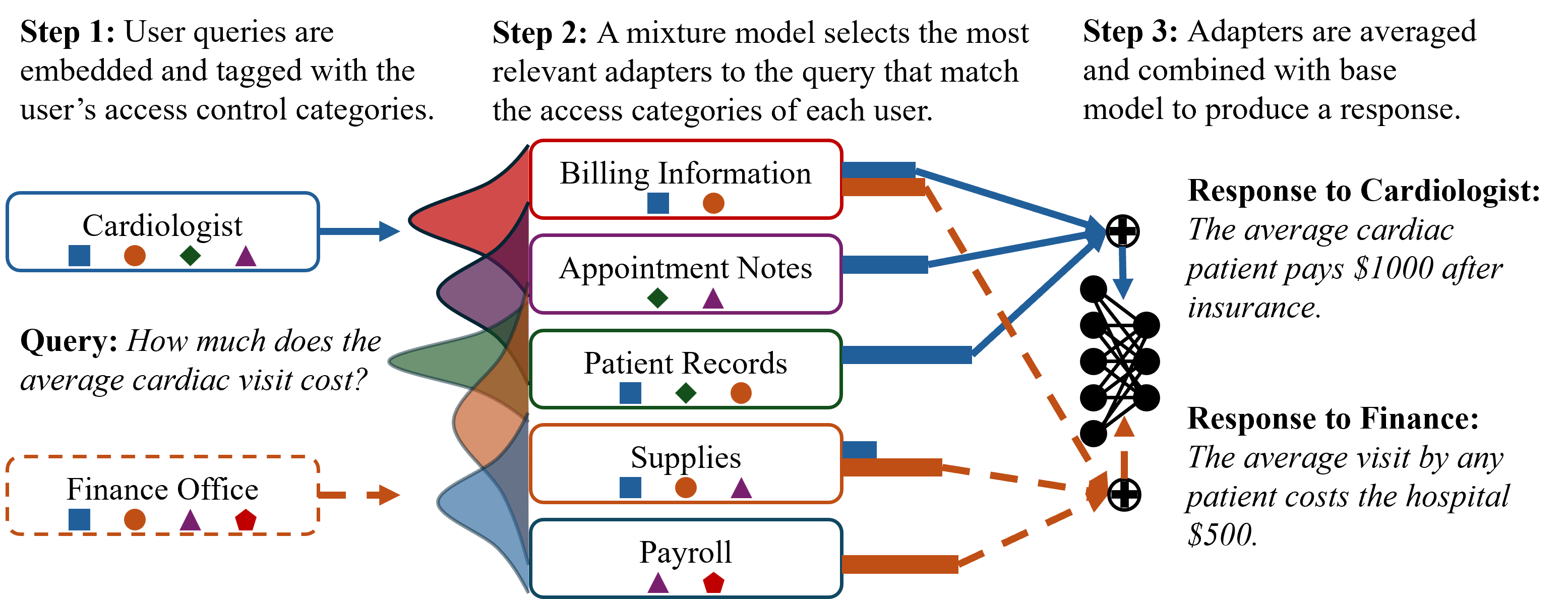}
    \caption{Motivating application of AdapterSwap. A mixture model selects the most relevant adapters to each users' query with the appropriate access-controls (indicated by shapes). Selected adapters are then combined and applied to a base model to produce personalized responses for each user.}
    \label{fig:adapter-swap}
\end{figure}

A motivating application of AdapterSwap is displayed in \autoref{fig:adapter-swap}. In this fictitious example, AdapterSwap has been fit to hospital data with subsets of the data subject to various access-controls. Both a cardiologist and member of the finance office submit the query `\emph{How much does the average cardiac visit cost?}' The retriever model selects the most relevant adapters to the query for which each user has access. The cardiologist's unique access to appointment notes and patient records enables the model to access specific payments from cardiac patients and respond accordingly. In contrast, the finance office's access to payroll and supply expenditures results in a response from the hospital's perspective without leaking the private patient information to the unauthorized employee. 

The rest of this paper is structured as follows. In \autoref{motivation} we discuss challenges which motivate our research. In \autoref{background} we provide context with prior works from which we build upon. \autoref{adapterswap} details AdapterSwap, our proposed approach. We demonstrate and quantify benefits of AdapterSwap through careful experimentation in \autoref{experiments}. We discuss additional related works in \autoref{related}. Finally, we conclude and suggest future research in \autoref{conclusion}.

Specifically, in this work we:
\begin{itemize}
     \item Develop an efficient approach for continuous knowledge acquisition through the training and dynamic composition of multiple LoRA adapters;
    \item Demonstrate our method's ability to guarantee data access-control and handle data removal in an efficient manner;
    \item Quantify our performance via a document completion task across a diverse set of LLMs: Falcon-7B, Gemma-7B, Llama-2-7B, and Mistral-7B; and
    \item Show that our approach mitigates forgetting better than iterative fine-tuning and retraining.
\end{itemize}

\section{Motivation}
\label{motivation}

Our work is primarily motivated by three issues faced when fine-tuning and deploying LLMs in real-world organizations. Namely, how to utilize data with access-controls, how to remove knowledge from a model retroactively, and how to update knowledge over time as new data becomes available. 

\subsection{Data Access-control}

It is common for organizations to apply access-control to their sensitive data \cite{hu2006assessment}. For example, only certain employees at a hospital have access to patient records to protect privacy. Similarly, employees at a law firm are concerned with attorney-client privilege. Access-control can also be an important aspect of business models such as a news aggregator giving access to users and their personalized chatbots based on their paid subscriptions \cite{tars}. We would like a language model trained on these data to inherit and guarantee these access restrictions. 

\subsection{Data Protection and Removal}

Organizations can unexpectedly lose the rights to maintain certain data. For example, data protection policies such as the General Data Protection Regulation (GDPR) \cite{GDPR2016a} allow users to recall their data from corporations. \emph{The Stack}, a popular dataset comprised of source code repositories, allows users to opt-out of having their code in future versions of the dataset but offers no solution for models trained on previous versions \cite{Kocetkov2022TheStack}. Similarly, the removal of training data later found to be copyright protected or unlicensed might be mandated through legal action, a growing concern for LM producers \cite{NYT2023, NYT2024}. Existing models provide no mechanism to remove all knowledge from individual training examples, so to comply with these mandates would require the entire model to be retrained. Therefore, we would like a more efficient approach to guarantee the removal of data from models. 

\subsection{Catastrophic Forgetting}

Organizations often have access to continuous or evolving streams of data. \emph{Catastrophic forgetting} is an issue that arises when a machine learning model forgets previously seen information as it learns from new data \cite{MCCLOSKEY1989109}. LLMs have been shown to suffer from forgetting during the fine-tuning process \cite{luo2023empirical}. We would like a method that addresses this issue and guarantees the ability to recall old information as new knowledge is continuously acquired.

\section{Background}
\label{background}

\subsection{Parameter Efficient Fine-Tuning}

As language models have become more specialized, growing in size and capabilities, fine-tuning an entire model has become unreasonable on commodity hardware. To address these challenges, several methods have been developed for performing \emph{parameter efficient fine-tuning}.

References \cite{guo-etal-2021-parameter} and \cite{NEURIPS2021_cb2653f5} present techniques for training a sparse subset of parameters in a multi-task setting. Alternatively, \emph{prompt tuning} and \emph{prefix tuning} concatenate learned task-specific embeddings to the sequence of inputs or activations being processed by a model \cite{lester-etal-2021-power, li-liang-2021-prefix}. 

In the wider context of transfer learning, adapter layers provide a straight forward mechanism to efficiently and effectively generalize a base model to a target task or domain by fine-tuning a new set of parameters (an adapter) on the target data \cite{bapna-firat-2019-simple, houlsby2019parameterefficient}. Low-rank adapters (LoRA) have emerged as a parameter efficient approach to fine-tuning large language models with reasonable amounts of compute
\cite{DBLP:journals/corr/abs-2106-09685}. In this work, we leverage LoRAs to continuously update and control a language model's knowledge.

\subsection{Model Averaging and Segmentation}

Several approaches have been suggested for combining model weights or outputs with demonstrated increases in performance or efficiency in certain scenarios. Reference \cite{pmlr-v162-wortsman22a} proposed \emph{Model Soups} which average the weights of multiple models trained on the same data with different hyper-parameters. While they show that the soups increase performance and robustness of language models, they do not address combinations of models trained on separate data. 

\emph{AdapterFusion} was introduced by \cite{pfeiffer-etal-2021-adapterfusion} as an approach to multi-task learning that segments task-specific knowledge into separate adapters that are then combined via an attention mechanism. While segmenting tasks is similar to segmenting data based on access controls, the attention mechanism adds additional complexity and the model dependency prevents the efficient removal of data. 

Reference \cite{ruckle-etal-2021-adapterdrop} extended ideas from AdapterFusion with their approach \emph{AdapterDrop}. AdapterDrop prunes adapters for increased efficiency but still lacks the ability to address efficient data removal or continual training.

\emph{AdaMix} was proposed by \cite{wang-etal-2022-adamix} and uses a mixture-of-experts approach to combining adapters at each layer for the purpose of parameter sharing but not for specific knowledge segmentation. 

\emph{Hierarchical Adapters} by \cite{chronopoulou-etal-2022-efficient} and \emph{AdapterSoup} by \cite{chronopoulou-etal-2023-adaptersoup} are the most similar to our approach as they train individual adapters on segmented domains in the training data, but their focus is on combining the adapters to perform well with out-of-domain queries, while our focus is on in-domain access-control and efficient knowledge deletion. 

\section{AdapterSwap}
\label{adapterswap}

\subsection{Data Segmentation and Adapter Training}
The first stage of our approach is choosing a data partitioning scheme. Data is segmented into separate access-control categories and further sharded based on desired \emph{per-shard} compute requirements\footnote{See \autoref{partitioning} for a discussion on this topic.}. A pretrained LLM is then used as a base model for fine-tuning a separate LoRA adapter per data partition. As the information from each partition is isolated to a single adapter, the adapter inherits the access-control categories of the data. This partitioning and fine-tuning stage can be done once for a static dataset, or on a continual basis as data arrives. 

\subsection{Retrieval Model}
Similar to AdapterSoup, we use a Gaussian Mixture Model (GMM) to retrieve the subset of adapters relevant to a given query during inference. To fit the GMM we use a pretrained SBERT \cite{reimers-gurevych-2019-sentence} model to embed randomly held-out samples from each partition into vectors of dimension 768\footnote{Specifically, we use the all-mpnet-base-v2 model from \url{https://huggingface.co/sentence-transformers/all-mpnet-base-v2.}}. We further reduce the dimension of these vectors by applying linear discriminant analysis (LDA). While previous approaches such as \cite{chronopoulou-etal-2023-adaptersoup} and \cite{aharoni-goldberg-2020-unsupervised} use principal components analysis (PCA), in this work we compare PCA to LDA. LDA has the theoretical benefit of maximizing the linear separability of the clusters in the lower dimensional space by using their labels in a supervised manner. We found that using LDA over PCA significantly improved our downstream retrieval accuracy which we discuss in \autoref{retieval}. Finally, the GMM is fit on the lower dimensional vectors with the number of components equal to the number of adapters. The efficiency of LDA and GMMs allows for cheap retraining of the retriever if new adapters are added over time.

\subsection{Inference}
During inference, queries are embedded using the same approach, and the GMM is used to rank potential adapters. The user's access control categories are applied so that restricted adapters are prevented from being selected. We explore several retrieval modes where either the top-1, top-2, or top-3 adapters with the highest GMM density are selected and combined with the base model. We also attempted averaging all non-restricted adapters with equal weight or by weighting according to GMM density, but the results for those scenarios were poor and are therefore omitted. In practice, the choice of retrieval mode could vary by context. For example, our experiments show that top-1 results are likely best conditioned on knowledge that the information being retrieved was isolated to a single adapter. However, combinations of multiple adapters have been shown to perform better for out of domain queries \cite{chronopoulou-etal-2023-adaptersoup}.

\subsection{Data Removal}
Finally, if circumstances arise that require permanently removing data from our training set, only the weights associated with the LoRA adapter trained on the removed data need to be retrained; less than 0.1\% of the base model's parameters in our experiments. This contrasts with emerging approaches such as \cite{pan2024lisa} for efficiently fine-tuning over the entire base model; which offers no savings as all fine-tuned parameters would require discarding if data is removed.

\begin{figure}[ht]
    \centering
    \includegraphics[scale=0.52]{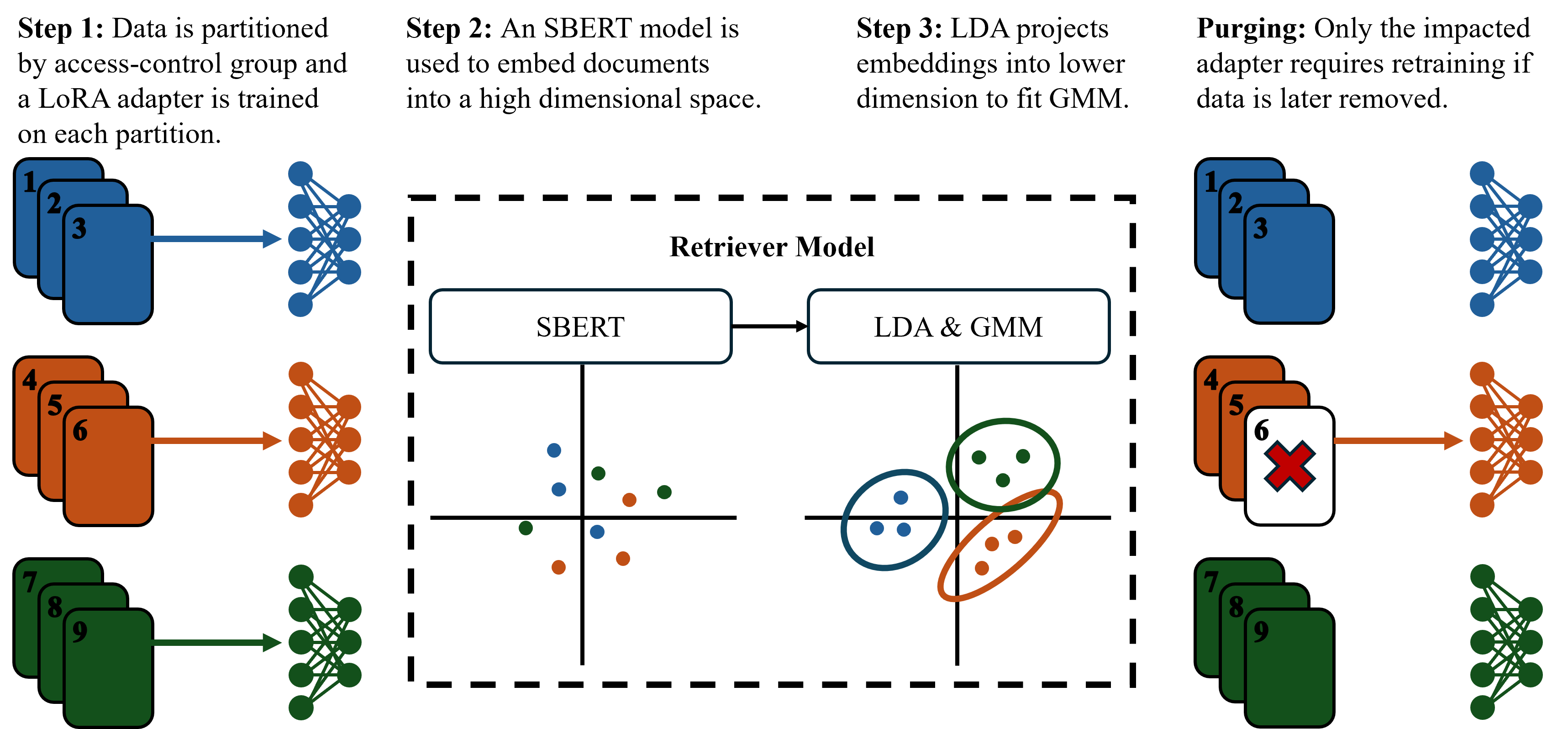}
    \caption{AdapterSwap overview. Individual adapters are trained on partioned access-control groups. A retriever model is fit using LDA and a GMM over SBERT representations. If data is removed only the impacted adapter requires retraining.}
    \label{fig:training}
\end{figure}

In \autoref{experiments} we demonstrate AdapterSwap using several models under both access-control and data removal scenarios. We also compare AdapterSwap's ability to prevent \emph{forgetting} with alternative methods for continuous learning. An overview of AdapterSwap training is illustrated in \autoref{fig:training}.

\section{Experiments}
\label{experiments}
We demonstrate the effectiveness by AdapterSwap through multiple experiments. First, we describe the datasets and models used for our experiments in \autoref{data_models}. In \autoref{partitioning} we establish the training times and baseline performance when fitting multiple adapters over a sharded dataset.

Next, in \autoref{retieval} we quantify our ability to retrieve and compose mixtures of adapters during inference. We demonstrate AdapterSwap's ability to conform to data access-controls in \autoref{access_control} and effectiveness when removing data in \autoref{purging}. Finally, we compare AdapterSwap's resilience to forgetting against two alternative approaches for continuous learning in \autoref{forgetting_results} 

For all of our experiments, we evaluate AdapterSwap by segmenting documents into equal halves and measuring the perplexity of the model while force decoding the second half given the first. This document completion task is suited for determining if a model has trained on and remembered particular samples from the datasets. We also report training times in GPU Hours using a single 80GB A100 GPU.

\subsection{Data and Models}
\label{data_models}
We use two datasets for our experimentation. First, we use the subset of C4 \cite{JMLR:v21:20-074} utilized by \cite{chronopoulou-etal-2023-adaptersoup} which contains 21 website domains where unique pages from the domain represent separate documents. We treat each domain as a separate LoRA training group for our experiments involving access-control and data purging. The list of training domains and their corresponding document counts are shown in \autoref{tab:c4_domains}.

We also use an English subset of the WMT News Crawl Dataset \cite{kocmi-etal-2022-findings}. Passages from articles published in the year 2020 were extracted and deduplicated following \cite{streamingqa2022}. These passages were then segmented into LoRA training groups based on their month of publication. The chronological nature of this dataset makes it suitable for measuring a model's ability to recall previous training data as subsequent months are trained.

We leverage a diverse set of LMs as base models to ensure our approach generalizes. We replicate experiments across Falcon-7B \cite{penedo2023refinedweb}, Gemma-7B \cite{Gemma}, Llama-2-7B \cite{Touvron2023Llama2O} and Mistral-7B-v0.1 \cite{jiang2023mistral}.

\begin{table}[ht]
\caption{Subset of domains from C4 used as AdapterSwap training groups. Size is represented by the number of documents from each domain.}
    \small
    \centering
    \begin{tabular}{l|c}
        \textbf{Domain} & \textbf{Size } \\
        \toprule
        androidheadlines.com & 41894\\
        booking.com & 57218\\
        csmonitor.com & 47625\\
        dailymail.co.uk & 77150\\
        entrepreneur.com & 39373\\
        eventbrite.com & 85647\\
        express.co.uk & 72435\\
        forums.macrumors.com & 68513\\
        frontiersin.org & 12053\\
        glassdoor.com & 40227\\
        ign.com & 41275\\
        insiderpages.com & 48072\\
        instructables.com & 72154\\
        journals.plos.org & 10630\\
        librarything.com & 41617\\
        link.springer.com & 82720\\
        lonelyplanet.com & 39284\\
        medium.com & 48522\\
        npr.org & 55632\\
        pcworld.com & 40202\\
        wired.com & 42061\\
    \end{tabular}
    \label{tab:c4_domains}
\end{table}

We utilize the HuggingFace \cite{wolf2020huggingfaces} library to access the pretrained models and fit LoRA adapters using Parameter Efficient Fine-Tuning \cite{peft}. All adapters were trained on a single 80GB A100 GPU to standardize timing comparisons. In practice, AdapterSwap can be trained in parallel across several devices. Each adapter was trained for 10 epochs with a batch size of 20. Rank 32 LoRA adapters were applied to all attention layers for the News Crawl data and rank 64 adapters on all linear layers for C4. All adapters were initialized with the same random seed, which was identified by \cite{chronopoulou-etal-2023-adaptersoup} as being necessary for adapter mixing. A detailed list of hyperparameters are included in appendix.

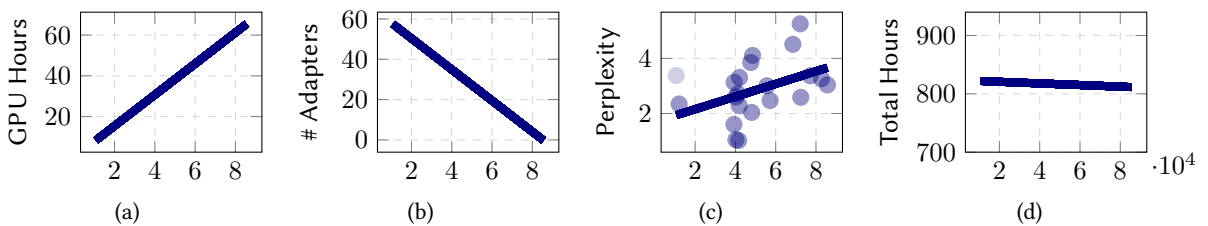
\begin{figure}[h]
    \centering
    \subfloat[]{
          \small
          \begin{tikzpicture}
          \begin{axis}[
            xtick scale label code/.code={},
              width=0.25\columnwidth,
              grid=major,
              grid style={dashed,gray!30},
              ylabel=GPU Hours,
              ]
            \addplot[line width=3pt, darkblue] table[x={size}, y={create col/linear regression={y={time}}}, col sep=comma]{results/perplexity_size_time.csv};
            \end{axis}    
        \end{tikzpicture}}\hfill
    \subfloat[]{
      \small
      \begin{tikzpicture}
      \begin{axis}[
        xtick scale label code/.code={},
          width=0.25\columnwidth,
          grid=major,
          grid style={dashed,gray!30},
          ylabel= \# Adapters,
          ]
        \addplot[line width=3pt, darkblue] table[x={size}, y={create col/linear regression={y={adapters}}}, col sep=comma]{results/number_buckets.csv};
         \end{axis}
    \end{tikzpicture}}\hfill
    \subfloat[]{
      \small
      \begin{tikzpicture}
      \begin{axis}[
        xtick scale label code/.code={},
          width=0.25\columnwidth,
          grid=major,
          grid style={dashed,gray!30},
          ylabel=Perplexity,
          ]
        \addplot[only marks, mark size=3pt, darkblue, opacity=0.1] table[x={size}, y={perplexity}, col sep=comma]{results/perplexity_size_time.csv};
        \addplot[line width=3pt, darkblue] table[x={size}, y={create col/linear regression={y={perplexity}}}, col sep=comma]{results/perplexity_size_time.csv};
        \end{axis}
        \end{tikzpicture}} \hfill
    \subfloat[]{
          \small
          \begin{tikzpicture}
          \begin{axis}[
            x tick scale label style={at={(1.15,0)}},
              width=0.25\columnwidth,
              grid=major,
              grid style={dashed,gray!30},
              ylabel=Total Hours,
              ymin=700,
              ymax=940
              ]
            \addplot[line width=3pt, darkblue] table[x={size}, y={create col/linear regression={y={total}}}, col sep=comma]{results/number_buckets.csv};
            \end{axis}
        \end{tikzpicture}}
    \caption{(a) Average training time for a single adapter given the data partition size. (b) Total number of adapters needed per data partition size for 1,064,304 documents divided equally among partitions. (c) Observed perplexity per partition size when partitioning dataset by domain. (d) Total GPU hours required to train all adapters using an equal partition size per adapter.}
    \label{fig:quad_plot}
\end{figure}

\subsection{Shard Size, Time, and Performance Trade-offs}
\label{partitioning}
For each model, we trained a separate adapter on all training groups. Because our groups naturally differ in size we are able to measure average training time and performance for different sharding strategies. \autoref{fig:quad_plot} displays differences in training time and document completion performance as a function of shard size. Partitioning the dataset into smaller groups results in the need for more adapters but enables faster training of each. The individual training time is an important characteristic when faced with the need to retrain adapters if data is later purged. We observe that smaller partition sizes tend to result in better perplexities, likely due to adapters having a higher ratio of parameters to training tokens. Overall, the total GPU hours is roughly equivalent across partitioning schemes, with a slight overhead resulting from adding each additional adapter. 

\subsection{Retrieval}
\label{retieval}

An optimal retriever should return the adapters which have knowledge related to a given query. In our case we know each sample was seen by only a single adapter, which we refer to as the \emph{oracle} adapter, but multiple adapters might be necessary in general information retrieval scenarios. \autoref{tab:retriever} displays the document completion performance across all models when using the oracle adapter, as well as an average of the top-1, top-2, and top-3 adapters as ranked by the retriever model. We also show the top-1 adapter with PCA as the projection method to compare with previous works \cite{chronopoulou-etal-2023-adaptersoup, aharoni-goldberg-2020-unsupervised}. For all models, using the top-1 adapter with LDA resulted in the best completions and retrieved the oracle adapter with accuracy varying from 69\% to 81\%. With the exception of the Gemma-7B model, results for the top-2 and top-3 schemes are reasonable and \cite{chronopoulou-etal-2023-adaptersoup} show that higher mixtures work well when the query is out of domain. The accuracy for retrieving the oracle adapter in the top-3 ranged from 93\% to 95\%, suggesting a potential inference scheme like \cite{feng2023cook}'s, where output is selected from the mixture and individual adapters via a separate content-selector. 

\begin{table}[ht]
\caption{Perplexity and accuracy retrieving oracle adapter (the adapter trained on the data being queried) when applying the top-3, top-2, and top-1 adapters from the retriever model. The oracle column corresponds to just using the single correct adapter for inference. We also compare top-1 performance when using PCA instead of LDA. A lower perplexity, and higher accuracy indicates better performance.}
\centering
\small
\begin{tabular}{lccccc}
\textbf{Model} &  $\downarrow(\uparrow)$\textbf{Top-3} &
$\downarrow(\uparrow)$\textbf{Top-2} &
$\downarrow(\uparrow)$\textbf{Top-1} &
$\downarrow(\uparrow)$\textbf{Top-1 PCA} &
$\downarrow$\textbf{Oracle} \\
\toprule
Falcon & 14.4 (93\%) & 14.4 (86\%) & \textbf{4.2} (77\%) & 5.4 (52\%) & 2.9 \\
Gemma & 1456 (93\%) & 588 (82\%)  & \textbf{3.1} (69\%) & 4.8 (46\%) & 1.7 \\
Llama-2 & 25.8 (95\%) & 19.5 (86\%) & \textbf{2.4} (81\%) & 3.3 (53\%) & 1.8 \\
Mistral & 43.0 (93\%) & 34.8 (88\%) & \textbf{1.9} (75\%) & 6.7 (45\%) & 1.4 \\
\end{tabular}

\label{tab:retriever}
\end{table}

\subsection{Access-Control}
\label{access_control}

AdapterSwap can be directly applied to scenarios where data is organized into access-control categories. We simulate this with the C4 dataset by assigning each domain to a different category. For each document completion, we use our retrieval model to return the top-1 adapter under two scenarios: with access to all adapters and with access to all but the adapter trained on the restricted domain. 

The results are summarized in \autoref{tab:access_control}. As expected, the best completion was achieved using the adapter with access to the relevant data, and performance dropped significantly when the retriever did not have access to the domain. This demonstrates AdapterSwap's ability to enforce access-control at the adapter level, providing the best results to users while simultaneously preventing unauthorized access to restricted training data.

\begin{table}[ht]
\caption{Perplexity across access-control scenarios. \emph{No Access}: Using the top adapter retrieved for the query excluding the oracle adapter. \emph{With Access}: Using the top adapter retrieved among all adapters.}
\centering
\small
\begin{tabular}{lccc}
\textbf{Model} & $\downarrow$\textbf{No Access} & $\downarrow$\textbf{With Access} \\
\toprule
Falcon & 16.1 & \textbf{4.2} \\
Gemma & 68.0 & \textbf{3.1} \\
Llama-2 & 15.2 & \textbf{2.4} \\
Mistral & 28.2 & \textbf{1.9} \\
\end{tabular}

\label{tab:access_control}
\end{table}

\subsection{Data Removal}
\label{purging}

We also measure our ability to efficiently purge documents from our dataset after training. When a piece of data is removed from the corpus, only the adapter fine-tuned on that data requires retraining. We show this by attempting to complete documents for each adapter before and after removing them and retraining the adapter. 

\autoref{tab:purge} summarizes the results for this experiment. We see a significant performance drop when trying to complete documents that have been purged from their adapter as the model is now guaranteed to have lost access to that data. Referring back to \autoref{fig:quad_plot}, retraining a single adapter is up to 80x more efficient than if you had to fine-tune over the entire dataset. Purging with AdapterSwap provides the guarantee that removed documents will not contribute to later inferences without the huge cost of full retraining.

\begin{table}[ht]
\caption{Perplexity completing documents using an adapter with (\emph{Before Purge}) and without (\emph{Purged}) the purged data.}
\centering
\small
\begin{tabular}{lcc}
\textbf{Model} & $\downarrow$\textbf{Before Purge}   & $\downarrow$\textbf{Purged} \\
\toprule
Falcon & \textbf{2.9} & 13.4 \\ 
Gemma & \textbf{1.7} & 25.8 \\
Llama-2 & \textbf{1.8} & 11.9 \\
Mistral & \textbf{1.4} & 15.4 \\
\end{tabular}
\label{tab:purge}
\end{table}

\subsection{Catastrophic Forgetting}
\label{forgetting_results}
Finally, we compare AdapterSwap's ability to recall past information with two alternative strategies. A naive approach to handling new streams of data is to iteratively fine-tune a LM as new data becomes available. This workflow can cause older information to be `overwritten’ within the architecture’s parameters. Alternatively, as new data is added, the model can be fine-tuned from scratch on all available data at once. While this mitigates some `forgetting' it requires longer training times as models are discarded and retrained. We use News Crawl to evaluate both methods and compare to AdapterSwap. We iteratively measure performance on the first month of our dataset as we add subsequent months of data to our models. For this experiment we only use Falcon-7B as the baseline due to the increased computational demand required by the alternative approaches. 

\begin{figure}[ht]
    \centering
    \begin{tikzpicture}
    \begin{axis}[
    width=0.35\columnwidth,
	xlabel=Months of Training,
	ylabel=Perplexity,
	grid=both,
    legend style={legend pos=north west,font=\tiny},
	no marks]
\addplot[line width=2pt,solid,color=darkblue] %
	table[x=months,y=finetuned,col sep=comma]{results/forgetting.csv};
\addlegendentry{FT};
\addplot[line width=2pt,solid,color=darkorange]
    table[x=months,y=retrained,col sep=comma]{results/forgetting.csv};
\addlegendentry{RT};
\addplot[line width=2pt,dashed,color=black]
    table[x=months,y=all_adapters,col sep=comma]{results/forgetting.csv};
\addlegendentry{AS};

\end{axis}
\end{tikzpicture}
    \caption{Perplexity of the first month of data measured after month by month training. FT indicates chronological fine-tuning as data becomes available. RT indicates retraining with new data and all preceding data. AS indicates AdapterSwap performance using the first month adapter.}
    \label{fig:forgetting}
\end{figure}
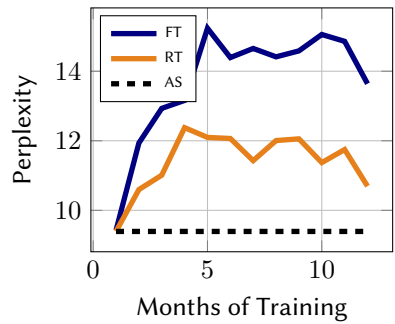

\autoref{fig:forgetting} displays the performance of AdapterSwap compared to chronological fine-tuning and full retraining. Chronological fine-tuning quickly degrades as more data is presented to the model. Retraining performs better, but suffers from the fixed capacity of a single adapter. AdapterSwap maintains a static performance when recalling data from the first month as that adapter remains unchanged over time.

\section{Additional Related Work}
\label{related}
\subsection{Knowledge Editing}
Recent work in knowledge editing of LLMs has considered approaches which either add additional parameters to a model, or directly edit existing parameters to update information \cite{yao-etal-2023-editing}. Directly updating the existing parameters is attractive as it does not require any additional parameters, and updates can be applied whenever new knowledge is available \cite{de-cao-etal-2021-editing, meng2022locating}. Knowledge vectors, in combination with hidden representations of specific entities, have also been proposed as a tool to update or remove knowledge \cite{hernandez2023inspecting}. In all cases, these approaches lack the ability to guarantee that any specific training example can be removed entirely from the model.

\subsection{Retrieval Augmented Generation}

Retrieval-Augmented Generation (RAG) has become a popular approach to incorporating new data into a pre-trained LLM without having to rely on retraining or fine-tuning \cite{lewis2021retrievalaugmented, gao2024retrievalaugmented}. While it offers some advantages, there remain challenges that can make deploying an effective RAG-based solution difficult. For example, RAG is limited by how much retrieved context can be employed based on the underlying LLM's context window size. This contrasts with AdapterSwap where each adapter in our experiments represented more than 50 million tokens on average. Recent work has shown that all evidence in the context window is not treated equally, with models favoring evidence at the start and end of each window \cite{Liu2023LostIT}. Further, RAG is fully dependent on the ability of a retriever model to locate all relevant documents without introducing too much noise into the context \cite{barnett2024seven, gao2024retrievalaugmented, RAGReyes}. In addition, as transformers are quadratic in the number of tokens considered, then requiring forced decoding over retrieved content can add significant latency at inference time. While our solution avoids these issues, we note that AdapterSwap does not preclude the use of RAG, and a hybrid approach could be useful in some circumstances.

\subsection{Federated Learning}
Federated learning \cite{mcmahan2017communication} also deals with learning from siloed data, typically aggregating gradients on local data before averaging into a global model. However, federated learning is intended to produce a single centralized model without mixing data silos and thus does not provide any mechanism for access control or deletion. Some work has combined federated learning and PEFT methods \cite{kim-etal-2023-client, babakniya2023slora, zhang-etal-2023-fedpetuning}, but these do not address adapter mixing, deletion, or data silos as distinct knowledge sources. Furthermore, the privacy benefits of federated learning are unclear when applied to LLMs with large capacity for memorization \cite{gupta2022recovering,shi2023detecting, tirumala2022memorization, carlini2023quantifying}.

\section{Conclusion}
\label{conclusion}

In this paper, we introduced AdapterSwap, a parameter efficient approach to continuous learning with access-control and data removal guarantees. We fine-tuned adapters using four modern pretrained language models on separate domains. We showed that knowledge from specific domains can be masked via access-control by preventing a retriever from accessing the controlled adapter at inference. The multiple-adapter scheme also enables efficient knowledge removal via data deletion and adapter retraining. The non-parametric behavior of AdapterSwap enables knowledge from the past to be retained, and we showed that AdapterSwap outperforms both chronological fine-tuning and retraining.

AdapterSwap enables a rich set of future research opportunities. We would like to directly improve the
approach by exploring better retrieval and adapter mixing methods. Additionally, AdapterSwap could
be further scaled to specific down-stream tasks such as question answering and directly compared or
combined with alternative data management schemes such as a RAG framework.

\bibliography{anthology, custom}

\appendix
\section{Training Details}
\label{hypers}

\begin{itemize}

\item We trained each adapter for our experiments on a single 80GB A100 GPU for 10 epochs with a batch size of 4 and 5 gradient accumulation steps. We utilized the AdamW optimizer with default settings.

\item For adapters trained on C4 domains we used rank 64 LoRAs with $\alpha = 128$ applied to all linear layers.

\item For the News Crawl experiment we used LoRA adapters with rank 32 and $\alpha = 64$ applied to just the attention layers.

\item For all experiments we initialized adapters using a random seed of 42. We confirm \cite{chronopoulou-etal-2023-adaptersoup}'s finding that using the same initialization is critical if mixing adapters.
\end{itemize}

\end{document}